%% file: main.tex
\documentclass{article} 
\usepackage{iclr2026_conference,times}

\input{math_commands.tex}

\usepackage[utf8]{inputenc} 
\usepackage[T1]{fontenc}    
\usepackage{hyperref}       
\usepackage{url}            
\usepackage{booktabs}       
\usepackage{amsfonts}       
\usepackage{wrapfig}        
\usepackage{subfig}         
\usepackage{nicefrac}       
\usepackage{microtype}      
\usepackage[dvipsnames]{xcolor}          
\usepackage{xpatch}
\usepackage{amsmath}
\usepackage{longtable}
\usepackage{amssymb}
\usepackage{pifont}
\usepackage{amsfonts}       
\usepackage{subcaption} 
\usepackage{siunitx}
\usepackage{pifont}
\usepackage{booktabs}
\usepackage{enumitem}
\usepackage{algorithm}
\usepackage[noend]{algpseudocode}
\usepackage{minted}
\usepackage{tikz,colortbl}

\usepackage{graphicx}
\usepackage[utf8]{inputenc}
\usepackage[T1]{fontenc}

\usepackage{amsfonts}
\usepackage{amsthm}
\usepackage{booktabs}
\usepackage{float}
\usepackage{algorithm}
\usepackage{subfig}
\usepackage{tikz}
\usepackage[edges]{forest}
\usepackage{multirow}
\usepackage{multicol}
\usepackage{tabularx}
\usepackage{soul}
\usepackage{eqnarray}
\usepackage{makecell}
\usepackage{comment}
\usepackage{array}
\usepackage{longtable}
\usepackage{amssymb}
\usepackage{pifont}
\usepackage{changepage}
\usepackage{amsthm}
\usepackage{xcolor}
\usepackage{mdframed}
\usepackage{xcolor}
\newmdenv[
  backgroundcolor=gray!10,   
  linecolor=gray!60,         
  linewidth=0.8pt,
  roundcorner=2pt,
  skipabove=6pt,
  skipbelow=6pt,
]{remarkbox}

\theoremstyle{plain}
\newtheorem{theorem}{Theorem}[section]

\newtheoremstyle{Astyle}
  {3pt}{3pt}
  {}{}
  {\bfseries}{}
  {0.5em}
  {(\thmname{#1}\thmnumber{#2})}

\theoremstyle{Astyle}
\newtheorem{assumption}{A}

\usepackage{amsmath}
\usepackage{amssymb}
\usepackage{mathtools}
\usepackage{amsthm}
\usepackage{multirow}
\usepackage{multicol}
\usepackage{tabularx}
\usepackage{graphicx}
\usepackage{xspace}

\usepackage[capitalize,noabbrev]{cleveref}

\theoremstyle{plain}

\newtheorem{lemma}[theorem]{Lemma}

\theoremstyle{definition}

\theoremstyle{remark}
\newtheorem{remark}[theorem]{Remark}

\newtheoremstyle{Astyle}
  {3pt}{3pt}
  {}{}
  {\bfseries}{}
  {0.5em}
  {(\thmname{#1}\thmnumber{#2})}

\theoremstyle{Astyle}

\newcommand{\para}[1]{\noindent\textbf{#1.}}

\newcommand{\sys}{\texttt{AttenMLP}\xspace}

\newcounter{RihanNOC}
\stepcounter{RihanNOC}

\usepackage{xpatch}
\makeatletter
\xapptocmd{\NAT@bibsetnum}{\setlength{\leftmargin}{0pt}\setlength{\itemindent}{\labelwidth}\addtolength{\itemindent}{\labelsep}}{}{}
\makeatother

\title{An Attention-based Feature Memory Design for Energy-Efficient Continual Learning}

\author{Yuandou Wang, Filip Gunnarsson, and  Rihan Hai
\\
Web Information Systems, EEMCS\\
Delft University of Technology\\
Delft, 2628 XE, The Netherlands  \\
}

\iclrfinalcopy 
\begin{document}

\maketitle

\begin{abstract}
Tabular data streams are increasingly prevalent in real-time decision-making across healthcare, finance, and the Internet of Things, often generated and processed on resource-constrained edge and mobile devices. Continual learning (CL) enables models to learn sequentially from such streams while retaining previously acquired knowledge. While recent CL advances have made significant progress in mitigating catastrophic forgetting, the energy and memory efficiency of CL for tabular data streams remains largely unexplored. To address this gap, we propose \sys, which integrates attention-based feature replay with context retrieval and sliding buffer updates within a minibatch training framework for streaming tabular learning.

We evaluate \sys against state-of-the-art (SOTA) tabular models on real-world concept drift benchmarks with temporal distribution shifts. Experimental results show that \sys achieves accuracy comparable to strong baselines without replay, while substantially reducing energy consumption through tunable design choices. In particular, with the proposed attention-based feature memory design, \sys costs a 0.062 decrease in final accuracy under the incremental concept drift dataset, while reducing energy usage up to 33.3\% compared to TabPFNv2. Under the abrupt concept drift dataset, \sys reduces 1.47\% energy consumption compared to TabR, at the cost of a 0.038 decrease in final accuracy. Although ranking third in global efficiency, \sys demonstrates energy-accuracy trade-offs across both abrupt and incremental concept drift scenarios compared to SOTA tabular models.
\end{abstract}

\input{chapters/S1_intro_v2}

\input{chapters/S2_related_work_v2}

\input{chapters/S3_method_v2}

\input{chapters/S4_energy_performance_analysis_v2}

\input{chapters/S5_experiments_v2}

\input{chapters/S6_conclusions_v2}

\bibliographystyle{iclr2026_conference}
\bibliography{Reference}

\newpage

\appendix
\end{document}

%% file: math_commands.tex
\usepackage{amsmath,amsfonts,bm}

\def\eqref#1{equation~\ref{#1}}

\def\1{\bm{1}}

\DeclareMathAlphabet{\mathsfit}{\encodingdefault}{\sfdefault}{m}{sl}
\SetMathAlphabet{\mathsfit}{bold}{\encodingdefault}{\sfdefault}{bx}{n}



%% file: chapters/S1_intro_v2.tex
\section{Introduction}
Tabular data, structured as a collection of features and instances, is one of the most common and practical data types in practical machine learning applications, for example, in both high-stakes domains and lower-stakes domains~\citep{amrollahi2022leveraging, ramjattan2024comparative, li2025unleashing}.
As such domains increasingly rely on streaming data sources, tabular data streams are gaining significant attention due to their ability to capture continuous, real-time updates rather than static snapshots~\citep{borisov2022deep}. In particular, most such scenarios often occur on edge devices, IoT systems, and mobile platforms, where energy budgets, battery life, and computational resources are severely constrained~\citep{chang2021survey}. 

To tackle those real-world dynamics, continual learning (CL)~\citep{wang2024comprehensive}, also referred to as lifelong learning~\citep{lee2020clinical}, enables models to incrementally acquire, update, accumulate, and exploit knowledge over time. 
While significant progress has been made on overcoming catastrophic forgetting~\citep{kemker2018measuring, li2019learn, bhat2022consistency} and knowledge transfer~\citep{ke2021achieving, li2024learning, shi2025continual}, much less is known about their computational analysis and energy efficiency~\citep{li2023large, Trinci2024}.

Energy-efficient continual learning (EECL) has become a practical necessity for real-world applications that require real-time adaptation on resource-constrained platforms~\citep{chavan2023towards, shi2024towards, Trinci2024, xiao2024hebbian}. Meanwhile, most CL progress to date targets image~\citep{Trinci2024,chavan2023towards,shi2024towards} and language tasks~\citep{li2025lsebmcl,wang2024comprehensive}. In contrast, tabular data streams remain underexplored. 
Tabular models that excel on static datasets do not transfer directly to non-stationary streams with tight memory, compute, and energy budgets. Existing CL methods rarely target these constraints. In particular, replay-based strategies rely on buffers that grow over time, increasing storage and compute, and hindering on-device deployment. This gap motivates methods for tabular streaming CL that sustain accuracy under distribution shift while operating at low energy cost, with fixed memory, and without storing raw examples. Moreover, trade-offs between energy consumption and predictive performance matter in lower-stakes domains, especially when the cost of electricity is taken into account. Achieving this under strict resource budgets while mitigating catastrophic forgetting remains a central challenge for Green AI~\citep{Henderson2020, Bouza2023, Trinci2024}. 

This paper introduces \sys, a novel EECL method for tabular streams, which integrates an attention-based feature memory for dynamic retrieval of historical embeddings, mitigating catastrophic forgetting. We utilize a multilayer perceptron (MLP) backbone to support this mechanism, leveraging its superior representation learning and high edge throughput. 
To be specific: 1) \sys employs a windowed scaled dot-product attention with a sliding feature buffer, enabling the model to adaptively attend to the most relevant parts of the stream while storing only latent features without needing to revisit raw historical data. 2) The resulting attended representation is concatenated and passed through two shared feed-forward layers followed by a classifier head, serving as the MLP learner for classification tasks. This design avoids the unbounded memory growth inherent to replay baselines~\citep{rebuffi2017icarl, li2017learning, lopez2017gradient}, while remaining computationally lightweight on resource-constrained devices. 
To evaluate energy-accuracy trade-offs in CL on tabular data streams, we provide quantitative Pareto status and global efficiency analysis. 


%% file: chapters/S2_related_work_v2.tex
\section{Related Work}

Traditional tabular data models can be roughly categorized into three main groups:  gradient-boosted decision trees (GBDTs), neural networks (NNs), and classic models. 

\para{GBDTs and their variants for CL} Traditional GBDTs such as XGBoost~\citep{chen2016xgboost}, LightGBM~\citep{ke2017lightgbm}, and CatBoost~\citep{prokhorenkova2018catboost} remain strong baselines for tabular classification due to their efficiency and robustness, especially on large or irregular static datasets. However, they are not naturally suited for CL: (1) new data typically requires retraining from scratch, since tree splits and boosting weights depend on the full dataset~\citep{chen2016xgboost, ke2017lightgbm, prokhorenkova2018catboost}; (2) without access to past data, models trained only on new samples overwrite previous knowledge, causing catastrophic forgetting~\citep{wang2024comprehensive}; and (3) unlike NNs, GBDTs lack mechanisms for knowledge transfer across tasks~\citep{ke2021achieving, parisi2019continual, de2021continual}. Extensions such as online bagging and boosting~\citep{oza2001online} or warm-starting~\citep{scikit-learn}, and adaptive XGBoost~\citep{montiel2020adaptive}, partially mitigate these issues, but remain limited in long-term knowledge retention due to the lack of representation reuse, especially when compared to neural CL methods.

\para{Classic models in CL} Both standard SVMs~\citep{cortes1995support} and decision trees~\citep{loh2011classification} are batch learners, requiring retraining on the full dataset when new tasks arrive. SVMs can be extended to CL through incremental or online variants such as incremental SVM~\citep{cauwenberghs2000incremental}, LASVM~\citep{bordes2005fast}, and NORMA~\citep{kivinen2004online}, which handle streaming updates but still face challenges with scalability, memory growth, and forgetting. k-NNs~\citep{cover1967nearest} trivially avoid forgetting if all data is stored, but this violates the constraint of no access to past raw inputs and is impractical under resource limits. Linear models~\citep{cox1958regression} are efficient but prone to forgetting under distribution shifts, as updates overwrite prior knowledge. Incremental decision trees, such as Hoeffding trees~\citep{domingos2000mining}, and streaming ensembles~\citep{bifet2010leveraging, gomes2017adaptive} can adapt to data streams without full retraining. Still, their accuracy degrades under severe drift, since they lack strong representation learning, and ensemble methods can be computationally expensive.

\para{Neural models in CL} Recent studies show that advanced NNs~\citep{zabergja2024deep, arik2021tabnet, kadra2021well, gorishniy2023tabr, hollmann2025accurate, ye2024modern, gorishniy2024tabm} can surpass GBDTs on static tabular data in certain regimes, e.g., with well-regularized MLPs~\citep{kadra2021well}, attention-based models such as SAINT~\citep{somepalli2021saint}, or meta-learned foundation models like TabPFN~\citep{hollmann2025accurate}. While their training or inference is typically more computationally intensive than that of GBDTs unless carefully tuned~\citep{kadra2021well}, NNs are generally better suited for streaming data, owing to their rich representations, incremental updates via stochastic gradient descent, and flexible architectures. However, vanilla NNs still suffer from catastrophic forgetting in the absence of CL strategies~\citep{wang2024comprehensive}. 

\para{CL strategies with neural models} In NNs, CL methods are commonly categorized into regularization-based approaches~\citep{kirkpatrick2017overcoming, zenke2017continual}, replay-based strategies~\citep{rebuffi2017icarl, Shin2017}, attention-based retrieval mechanisms~\citep{xu2019bayesian}, and architectural methods~\citep{rusu2016progressive}. Regularization-based methods, such as EWC~\citep{kirkpatrick2017overcoming}, SI~\citep{zenke2017continual}, MAS~\citep{aljundi2018memory}, A-GEM~\citep{chaudhry2018efficient}, and LwF~\citep{li2017learning}, mitigate forgetting by constraining updates to parameters deemed important for previously learned tasks. Replay-based strategies, including iCaRL~\citep{rebuffi2017icarl} and generative replay~\citep{Shin2017}, maintain past knowledge by rehearsing stored samples or synthetic data. Attention-based retrieval mechanisms, such as attentive experience replay~\citep{xu2019bayesian}, utilize an attention mechanism to leverage previous knowledge. Architectural methods, exemplified by PNNs~\citep{rusu2016progressive}, expand model capacity by freezing previously trained components and introducing new modules for incoming tasks.

Despite recent progress, EECL for tabular data streams remains largely unexplored~\citep{chavan2023towards, Trinci2024}. Real-world tables frequently undergo domain drift (e.g., quarterly finance transactions, evolving sensor logs, healthcare data) without changes to the label space. Yet, no standardized domain-incremental learning benchmark that considers energy-performance trade-offs currently exists for tabular streams. Moreover, pre-trained transformers for tabular data~\citep{gorishniy2021revisiting, hollmann2025accurate} and feature-level or attention-based CL strategies~\citep{vaswani2017attention, li2025lsebmcl, somepalli2021saint} show promise for low-storage, privacy-preserving CL, but their effectiveness under concept drifts has not been systematically evaluated. Here, we bridge this gap by introducing our method, establishing fair comparisons, and quantifying energy–performance trade-offs.

%% file: chapters/S3_method_v2.tex
\section{AttenMLP for EECL}


Owing to the general difficulty and diversity of challenges in CL, we focus on a simplified task incremental learning setting~\citep{parisi2019continual, de2021continual}. In this setting, a model is trained on a sequence of tasks $\{\mathcal{T}_t\}_{t=1}^T$, where the data for each task arrives incrementally at time $t$. 

\subsection{Problem Statement} 
Given a sequence of tasks $\{\mathcal{T}_t\}_{t=0}^T$ arriving incrementally, each task $\mathcal{T}_t$ consists of data $(\mathcal{X}_{t}, \mathcal{Y}_{t})$ sampled from a distribution $\mathcal{D}_t$. $\mathcal{X}_t = \{x_{t,i}\}_{i=1}^{n_t}$ represents the input features and $\mathcal{Y}_t = \{y_{t,i}\}_{i=1}^{n_t}$ denotes the corresponding ground truth labels. The goal is to learn under concept drift~\citep{souza2020challenges, hoens2012learning}, where the relationship between $\mathcal{X}_t$ and $\mathcal{X}_{t+1}$ evolves, i.e., $\mathcal{D}_t \neq \mathcal{D}_{t+1}$. 

We aim to design an incremental learner $f_{\theta}$ that updates online and minimizes the expected risk $\hat{L}_{t}(\theta)$ across all observed tasks, with no access to the data from earlier tasks, 
\begin{equation}
    \hat{L}_{t}(\theta) := \sum_{t=0}^{T}\mathbb{E}_{(\mathcal{X}_t, \mathcal{Y}_t) \sim \mathcal{D}_t}[\ell_{t}(\theta)], \label{eq:loss}
\end{equation}
where $\ell_t(\theta)$ represents the loss function of the incremental model $f_{\theta}$ with input $\mathcal{D}_t$, parameter $\theta$ and the historical information at time $t$. Additionally, we aim to achieve energy efficiency.

We formalize this with standard non-convex optimization assumptions. 
\begin{assumption}\label{A1: Bounded inputs.}
\textit{There exists $R_{\mathcal{X}} > 0$ such that for all $t, i$ and all samples $x_{t,i} \in \mathcal{X}_{t}$, we have $\|x_{t,i}\|_2 \le R_{\mathcal{X}}$.}
\end{assumption}

\begin{assumption}\label{A2: Bounded latent features.}
    \textit{The precomputed latent features are $\ell_2$-normalized, i.e., $\|h_{t,j}\|_2 \le 1$ for all $t,j$.}
\end{assumption}
\begin{assumption}\label{A3: Bounded parameters.}
    \textit{Training is performed with weight decay and early stopping, so that for some $R_\theta>0$, $\|\theta\|_2 \le R_\theta$ throughout optimization.}
\end{assumption}

\subsection{Architecture Overview}
\begin{figure}[!htb]
    \centering
    \includegraphics[width=.98\linewidth]{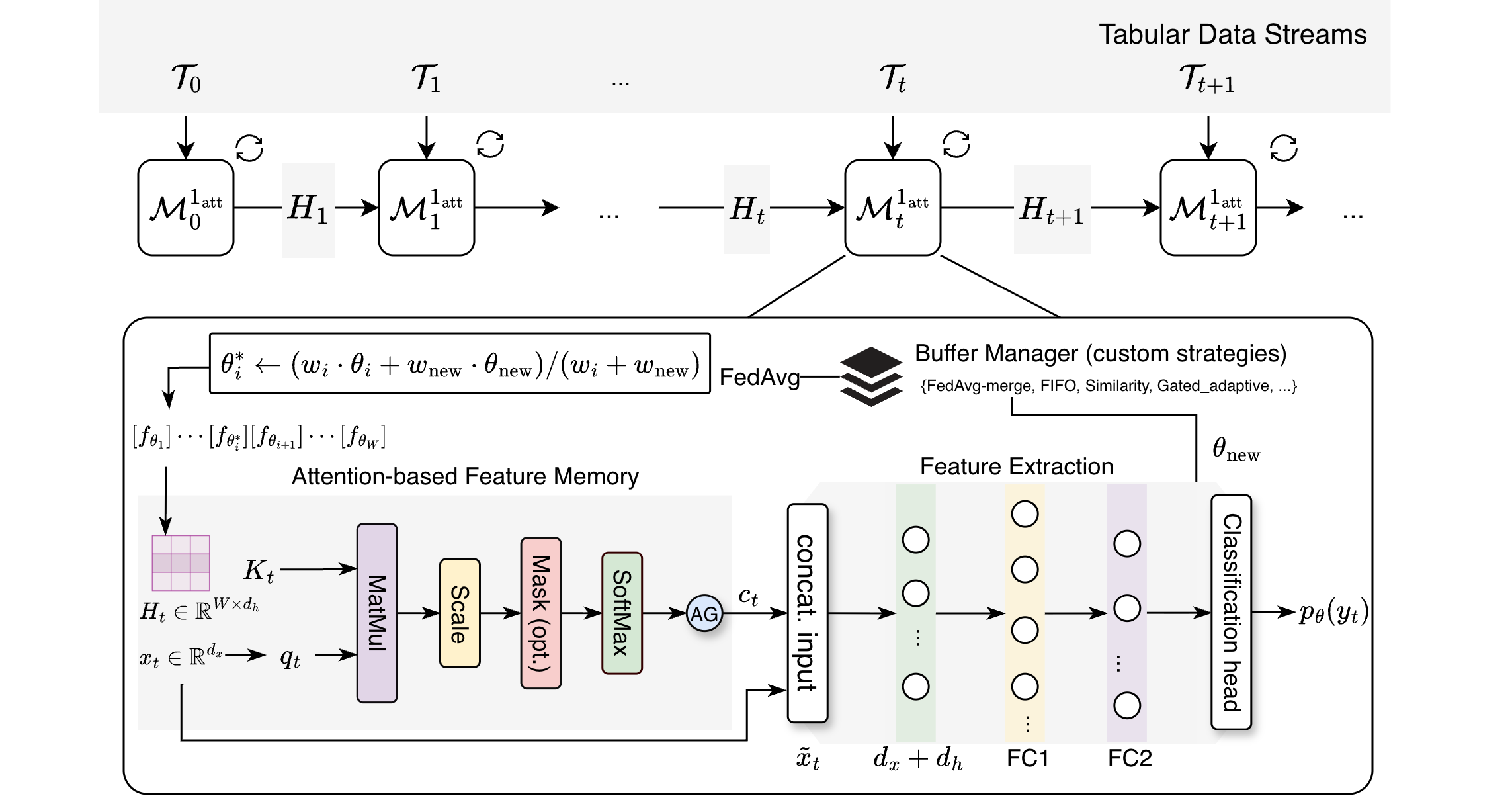}
    \caption{\sys architecture with custom buffer strategies. \sys sequentially takes $\mathcal{T}_t$ as raw input and outputs predictive performance $p_{\theta}(y_t)$. 
    }
    \label{fig: flowchart}
 \end{figure}

For efficient learning from the current task while maintaining performance on previously learned tasks, we propose the \sys architecture, as shown in Figure~\ref{fig: flowchart}. 

We employ two strategies: (1) processing each task with an augmented MLP learner module $\mathcal{M}^{1_{\text{att}}}$ that incorporates limited historical context in a window size through a variant of scaled dot-product attention; and (2) maintaining the windowed historical model buffer with a fixed memory over time to handle the concept drifts and energy use. 

Given the current input sample $x_t \in \mathbb{R}^{d_x}$, the size of the latent feature representation $d_h$ learned by the model, and a window size $W$, we denote
$H_t = [h_{t,1}, \dots, h_{t,W}]^\top \in \mathbb{R}^{W \times d_h}$
as the matrix, stacking the latent features in the window $W$ at time $t$. $h_{t,j} \in \mathbb{R}^{d_h}$ is the feature vector of the $j$-th most recent sample before $x_t$ (with $j=1$ being the most recent), precomputed from the buffer strategies via the historical learned models.
With learnable maps $W_q \in \mathbb{R}^{d_h \times d_x}$ and $W_k \in \mathbb{R}^{d_h \times d_h}$ (and biases $b_q \in \mathbb{R}^{d_h}$, $b_k \in \mathbb{R}^{d_h}$), \sys forms
\begin{align}
    q_t &= W_q x_t + b_q \in \mathbb{R}^{d_h}, \\
    K_t &= H_t W_k^\top + \mathbf{1}_W b_k^\top \in \mathbb{R}^{W \times d_h}, \\
    s_t &= K_t q_t \in \mathbb{R}^{W}, \\
    \tilde{s}_t &= \frac{1}{\sqrt{d_h}}\, s_t, \\
    \alpha_t &= \operatorname{softmax}(\tilde{s}_t) \in \mathbb{R}^{W}, \\
    c_t &= \alpha_t^\top K_t \in \mathbb{R}^{d_h}, 
\end{align}
where $\mathbf{1}_W \in \mathbb{R}^W$ is the all-ones vector used to broadcast the bias $b_k$ to all $W$ rows.
Here, $K_{t,j}\in \mathbb{R}^{d_h}$ is the $j$-th row of the corresponding key in the window. When the attention module is enabled, \sys updates $h_{t,j}$ from the historical model snapshots with custom buffer strategies. 
 
This work considers four custom buffer strategies: a) first-in-first-out (FIFO), which performs unconditional replacement in arrival order; b) a similarity-based strategy that enforces diversity by removing the most similar buffered feature; c) a gated-adaptive strategy that conditionally replaces features based on an adaptive similarity threshold; and d) the FedAvg-merge strategy. It maintains the fixed-size buffer of evolving model prototypes by merging each new model snapshot into its most similar stored model via weighted federated averaging~\citep{mcmahan2017communication}, rather than storing it separately.

The attention-based feature memory context is computed as a weighted sum of the keys, $c_t = \alpha_t^\top K_t \in \mathbb{R}^{d_h}$.
\sys then concatenates the context with the current input, $\tilde{x}_t = [x_t; c_t] \in \mathbb{R}^{d_x + d_h}$, and feeds it to the feature extractor to obtain the latent representation $f_t = \phi_{\theta}(\tilde{x}_t) \in \mathbb{R}^{d_h}$, where $\phi_{\theta}(\cdot)$ is a two-layer MLP.
Finally, the classifier outputs logits
$\hat{y}_t = W_{\mathrm{cls}} f_t + b_{\mathrm{cls}}$, where $W_{\mathrm{cls}} \in \mathbb{R}^{C \times d_h}$ and $b_{\mathrm{cls}} \in \mathbb{R}^{C}$.
The predictive distribution is given by $p_{\theta}(y_t) = \mathrm{softmax}(\hat{y}_t)$, in which $y_t \in \{1,\cdots, C\}$.

In the following sections, we detail the properties of the proposed attention-based feature memory design for achieving EECL over tabular data streams.

\subsection{Attention-based Feature Memory and Time Complexity}

Unlike replay buffers that grow with the number of seen samples, our attention-based feature memory has constant memory complexity over time. 
At time $t$, the buffer stores a maximum $W$ latent feature vectors of dimension $d_h$, i.e., $H_t\in \mathbb{R}^{W\times d_h}$, which results in $\mathcal{O}(Wd_h)$ memory per stream, independent of the stream length. In the batched implementation, each buffer holds $W$ feature tensors, each of shape $[B, d_h]$; hence, the runtime memory overhead is bounded by $\mathcal{O}(BWd_h)$. 
 
For a full forward pass with batch size $B$, the total computational efficiency of \sys is the sum of (i) the query projection $\mathcal{O}(Bd_{x}d_h)$, (ii) the key projection over the window $\mathcal{O}(BWd_h^2)$, (iii) attention score computation and aggregation $\mathcal{O}(BWd_h+BW+BWd_h)$, as well as (iv) the remaining MLP, consisting of two fully-connected layers FC1: ($d_x+d_h$)$\rightarrow 512$ and FC2: $512\rightarrow d_h$, plus the classifier: $d_h\rightarrow C$, which costs $\mathcal{O}(B(d_x+d_h)\cdot512)+Bd_h\cdot512+Bd_hC)$. Overall, the forward-pass complexity is absorbed by $\mathcal{O}(Bd_h(d_{x}+W d_h))$, 
where the attention time grows linearly with $W$ and quadratically with $d_h$, due to the $d_h^2$ key layer.

Therefore, the attention-based feature memory introduces two linear projections (query and key), adding $\mathcal{O}(d_xd_h+d_h^2)$ parameters. The per-step attention computation scales as $\mathcal{O}(Bd_{x}d_h +BW d_h^2+BWd_h)$. At runtime, maintaining a sliding window of size $W$ incurs $\mathcal{O}(BWd_h)$ memory to store the buffered features in batched form. 
Since the buffer size is capped by $W$ and $d_h$, this yields constant memory over time with respect to the stream length.

\subsection{Convergence Analysis of AttenMLP}

Let $\mathcal{A}(x_t,H_t;\theta_{\mathrm{att}}):= c_t$ denote the attention-based context, where $\theta_{\mathrm{att}} = (W_q,b_q,W_k,b_k)$ and $H_t$ collects the past latent features at time $t$ (with $c_t=\mathbf{0}$ if the window is empty).
\begin{lemma}[Bounded context vector]
\label{lem:bounded-context-main}
Under (A\ref{A2: Bounded latent features.}) and (A\ref{A3: Bounded parameters.}), there exists $B_c > 0$ such that $\|c_t\|_2 \le B_c \text{ for all } t$.
\end{lemma}


\begin{lemma}[Smooth attention map]
\label{lem:att-smooth-main}
Under (A\ref{A1: Bounded inputs.})--(A\ref{A3: Bounded parameters.}), the map
$(x_t,H_t,\theta_{\mathrm{att}})\mapsto \mathcal{A}(x_t,H_t;\theta_{\mathrm{att}})$ is continuously differentiable, and its Jacobian with respect to $\theta_{\mathrm{att}}$ is bounded on the compact set $\mathcal{K} := \{(x_t,H_t,\theta_{\mathrm{att}}) : \|x_t\|_2 \le R_{\mathcal{X}},\ \|h_{t,j}\|_2 \le 1,\ \|\theta_{\mathrm{att}}\|_2 \le R_\theta\}$. 
In particular, there exists $L_{\mathrm{att}}>0$ such that for all $(x_t,H_t)$ satisfying the above bounds and all
$\theta_{\mathrm{att}}^{(1)},\theta_{\mathrm{att}}^{(2)}$ with $\|\theta_{\mathrm{att}}^{(1)}\|_2,\|\theta_{\mathrm{att}}^{(2)}\|_2\le R_\theta$, $\|\mathcal{A}(x_t,H_t;\theta_{\mathrm{att}}^{(1)})-\mathcal{A}(x_t,H_t;\theta_{\mathrm{att}}^{(2)})\|_2
\le L_{\mathrm{att}}\|\theta_{\mathrm{att}}^{(1)}-\theta_{\mathrm{att}}^{(2)}\|_2.$
\end{lemma}


Correspondingly, the full network (logits) for a single sample can be formalized as
\begin{align}
    f_\theta(x_t,H_t)
    := W_{\mathrm{cls}} \, \phi_\theta\bigl([x_t;\mathcal{A}(x_t,H_t;\theta_{\mathrm{att}})]\bigr) + b_{\mathrm{cls}},
\end{align}
where $\phi_\theta$ is the two-layer ReLU feature extractor. The per-sample cross-entropy loss is stated by
\begin{align}
    \ell_t(\theta) = \mathrm{CE}\bigl(f_\theta(x_t,H_t), y_t\bigr).
\end{align}




{
}

\begin{theorem}[Segment-wise regularity of empirical loss]
\label{thm:segment-loss-main}
Under (A\ref{A1: Bounded inputs.})--(A\ref{A3: Bounded parameters.}), the empirical loss $\hat{L}_t(\theta)$ is
(i) bounded below; and
(ii) locally Lipschitz on the compact set $\{\theta:\|\theta\|_2 \le R_\theta\}$ and differentiable almost everywhere on this set.
Moreover, wherever $\nabla \hat{L}_t(\theta)$ exists, it is bounded on $\{\theta:\|\theta\|_2 \le R_\theta\}$.
\end{theorem}


\begin{remark}
    For each fixed segment $\mathcal{T}_t$, the attention-based feature memory yields a bounded mapping from a finite latent window, so the per-segment optimization problem is well-behaved for first-order methods under standard assumptions.
Still, we do not claim convergence to a single global limit when the data distribution is non-stationary across $t$.
\end{remark}

%% file: chapters/S4_energy_performance_analysis_v2.tex
\subsection{Energy Efficiency Analysis of AttenMLP}
Consider that we use the attention-based feature memory in both training and inference.

For a fixed device and runtime configuration, we model the energy consumed per training step as an affine function of the step’s floating-point operations (FLOPs) $F_{\text{step}}$, plus a bounded additive overhead $E_0$~\citep{choi2013roofline, douwes2024computation, newkirk2025empirically}, i.e., 
\begin{align}
    E_{\text{step}} = \eta F_{\text{step}}+E_0,
\end{align}
where $\eta$ is a device-dependent energy-per-FLOP constant. 
We adopt the following assumptions:
\begin{assumption}\label{A4:energy}
  \textit{On a fixed hardware platform (GPU/CPU) and under a fixed runtime configuration, there exist constants $0 < \eta_{\min} \leq \eta_{\max}$ such that the energy-per-FLOP coefficient $\eta$ satisfies $\eta \in [\eta_{\min}, \eta_{\max}]$.} 
  Equivalently, the FLOP-dependent part of the per-step energy is bounded as $\eta_{\min} F_{\text{step}} \leq E_{\text{step}} - E_0 \leq \eta_{\max}F_{\text{step}}$.
\end{assumption}
\begin{assumption}\label{A5:energy}
    \textit{The additional system overhead per training step (e.g., warm-up overhead and cold-start data retrieval) is bounded by a constant $E_0$ that is independent of the sample index and segment.}
\end{assumption}


\begin{lemma}[FLOP complexity per sample]\label{lem:FLOP-complexity-per-sample}
For a single sample $x_{t,i}$, the FLOP count of one forward-and-backward training step of \sys satisfies $F_{\text{train}}(x_{t,i})\le K_{\text{arch}}(d_x d_h + W d_h^2+ W d_h+(d_x+d_h)\cdot 512+512\cdot d_h+d_h C)$, 
for an architecture-dependent constant $K_{\text{arch}}>0$ 
that is independent of $t$ and $n_t$. 
To simplify the expression, we drop the tiny pieces $Wd_h+W$ of the attention score computation and aggregation.
\end{lemma}

\begin{theorem}[Per-segment energy complexity bound]\label{theorem:energy-complexity-bound}
Consider a segment $\mathcal{T}_t$ with $n_t$ training samples. Under (A\ref{A4:energy})--(A\ref{A5:energy}) and Lemma~\ref{lem:FLOP-complexity-per-sample}, the total training energy $E^{\text{train}}_t$ consumed by \sys on this segment satisfies
\begin{align}
         E^{\text{train}}_t \le n_t\Bigl(\eta_{\max}\,F_{\text{train}}(x_{t,i}) + E_0^{\mathrm{train}}\Bigr) 
\end{align}
Similarly, if $F_{\text{infer}}$ denotes the per-sample inference FLOPs (forward pass only $K_{\mathrm{arch}}^{\mathrm{(fwd)}}<K_{\mathrm{arch}}$), then the inference energy $E^{\mathrm{infer}}_t$ on a test set of size $n_t^{\text{test}}$ admits 
\begin{align}
         E^{\mathrm{infer}}_t \le n_t^{\mathrm{test}}\Bigl(\eta_{\max}\,F_{\mathrm{infer}}(x_{t,i}) + E_0^{\mathrm{infer}}\Bigr),
\end{align}
for a inference overhead constant \(E_0^{\text{infer}}\ge 0\).
\end{theorem}

\begin{remark}
    The bounds above help interpret our empirical observations in two ways. First, on a fixed hardware platform and runtime configuration, the energy consumed by \sys grows at most linearly with the total number of processed examples (and, for a fixed number of training epochs, linearly in the number of training steps), resulting in a predictable energy profile. Second, the attention-based feature memory introduces a controlled additional cost dominated by the key projection term $W d_h^2$, which remains tunable in our implementation since $W$ and $d_h$ are configurable. Consequently, the measured energy (in Joules) is consistent with an energy complexity that is mostly linear in stream size. 
\end{remark}

\section{Energy-Accuracy Trade-offs}
In many optimization problems, objectives are inherently conflicting; for instance, improving the accuracy of an NN increases energy consumption or latency. 
A classical way to study such trade-offs is through Pareto front analysis~\citep{giagkiozis2014pareto}. 

Our convergence and energy bounds naturally lead to a bi-objective viewpoint, where we jointly consider the online learner's predictive performance and energy consumption.
For a fixed segment $\mathcal{T}_t$, an \sys configuration is often determined by its
architecture parameters $B, d_h, W$ and optimization budget (e.g., number of iterations, learning-rate schedule). Each such configuration yields a pair $\bigl(E_t(\theta), P_t(\theta)\bigr)$, where $E_t(\theta)$ denotes the total energy consumed at time $t$ and $P_t(\theta)$ denotes the resulting segment-wise model performance (e.g., balanced accuracy).

We say that a configuration $\theta^{(1)}$ \emph{Pareto-dominates}
$\theta^{(2)}$ if
\begin{align}
    E_t(\theta^{(1)}) \le E_t(\theta^{(2)}), \quad
    P_t(\theta^{(1)}) \ge P_t(\theta^{(2)}),
\end{align}
with at least one strict inequality. The \emph{Pareto set} $\mathcal{P}_t:= \bigl\{
        \theta : \nexists\, \theta' \ \text{s.t.}\
        E_t(\theta') \le E_t(\theta),\
        P_t(\theta') \ge P_t(\theta) 
    \bigr\}$, 
which collects all Pareto-efficient configurations, and its image in the energy-performance plane forms the \emph{Pareto frontier}. To quantitively measure the trade-offs, we define an efficiency metric as the normalized Euclidean distance to the ideal performance–energy operating point, which is given by
\begin{align}
    \mathrm{Efficiency} = 1 - \frac{\sqrt{(1-P(\theta))^2 + (\tilde{E}(\theta))^2}}{\sqrt{2}}, 
\end{align}
where $\tilde{E}(\theta)\in [0,1]$ denotes the normalized energy consumption. This metric jointly captures predictive performance and energy consumption, rewarding models that achieve a balanced trade-off between accuracy and energy consumption.

Intuitively, on each segment $\mathcal{T}_t$, our smoothness and Lipschitz-gradient assumptions imply that stochastic first-order methods require on the order of $1/\varepsilon^2$ iterations to reach an $\varepsilon$-stationary point of the empirical loss $\hat L_t(\theta)$ in expectation. Each iteration has a bounded computational cost proportional to $d_{x} d_h + W d_h^2 + W d_h+ (d_x+2d_h)512 + d_h C$ (Lemma~\ref{lem:FLOP-complexity-per-sample}), and our energy model (Theorem~\ref{theorem:energy-complexity-bound}) shows that energy is proportional to this cost up to device-dependent constants. 

%% file: chapters/S5_experiments_v2.tex
\section{Experiments}\label{sec:exp}

\para{Setup and Configuration} We conducted extensive experiments on an Alienware desktop equipped with Intel(R) Core(TM) i9-10900KF CPU@3.70GHz CPU model, 32GiB Comet Lake PCH Shared SRAM memory, and one NVIDIA GeForce RTX 2080 Ti GPU model. To measure energy consumption, we instrument our CL pipeline with the Python package codecarbon~\citep{benoitcourty202411171501} that estimates and tracks carbon emissions from the device.

\para{Datasets and Baselines} We evaluate our method on real-world concept drift benchmarks using River’s \textit{INSECTS} datasets\footnote{\url{https://riverml.xyz/dev/api/datasets/Insects/}}
, which are chosen to represent challenging drift scenarios with different patterns~\citep{souza2020challenges}. We consider both \textit{abrupt} and \textit{incremental} concept drift settings to assess model robustness under different drift dynamics.
We compare our {\sys} model against a comprehensive set of nine recent SOTA models for tabular classification, covering three distinct model categories: 1) {foundation models:} $\texttt{TabPFNv2}$~\citep{hollmann2025accurate}; 2) {deep NN baselines:} $\texttt{RealMLP}$~\citep{holzmuller2024better}, $\texttt{ModernNCA}$~\citep{ye2024modern},
$\texttt{TabM}$~\citep{gorishniy2024tabm},
$\texttt{TabR}$~\citep{gorishniy2023tabr},
and $\texttt{MLP}$~\citep{taud2017multilayer}; and 3) {GBDTs:} $\texttt{CatBoost}$~\citep{prokhorenkova2018catboost}, $\texttt{XGBoost}$~\citep{chen2016xgboost}, and $\texttt{LightGBM}$~\citep{ke2017lightgbm}. 

\para{Evaluation Protocol} A crucial consideration for the evaluation is that our selected baselines were primarily developed for static, independent, and identically distributed data. While an ideal comparison would involve creating a dedicated CL variant of each baseline, e.g., equipped with specialized components for memory and catastrophic forgetting mitigation, such an undertaking is outside the scope of this work. To establish a methodologically sound comparison, we standardize the data flow and retrain the baseline models per arriving segment without using attention-based feature memory context. With the segmental mode, each segment is split into 85\% training and 15\% evaluation using a stratified split when feasible. The training epoch and learning rate are set to $\mathrm{epochs}=500, lr=0.001$, and batch size is 256.
All models are evaluated using six metrics: balanced accuracy, log-loss, total energy consumed, and execution time. All reported statistics are computed from the experimentally collected results with a random seed of 42.

\subsection{Ablation Study: Impact of Attention, \protect{$d_h$, $W$}\protect, and Buffer Strategy Choices}
\para{Attention and buffer strategy choices} We first ablated the core components, i.e., the attention module and custom buffer strategies, on the \textit{abrupt-concept-drift} dataset to evaluate their impacts on \sys performance and energy consumption. 

\begin{wrapfigure}{r}{0.4\linewidth}
    \centering
    \includegraphics[width=\linewidth]{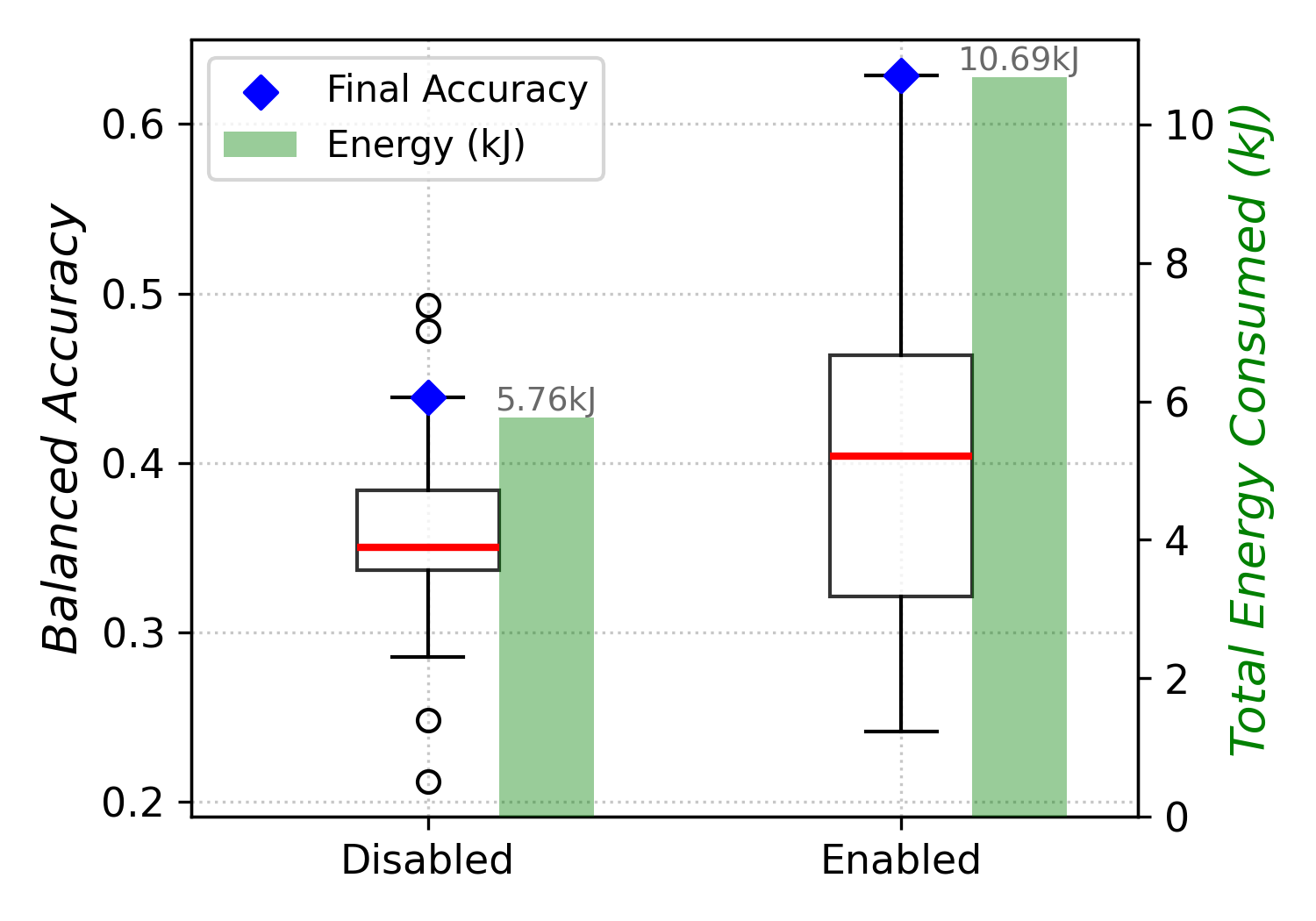}
    \caption{Attention usage comparison.}
    \label{fig: ablation1-atten}
    \centering
    \includegraphics[width=\linewidth]{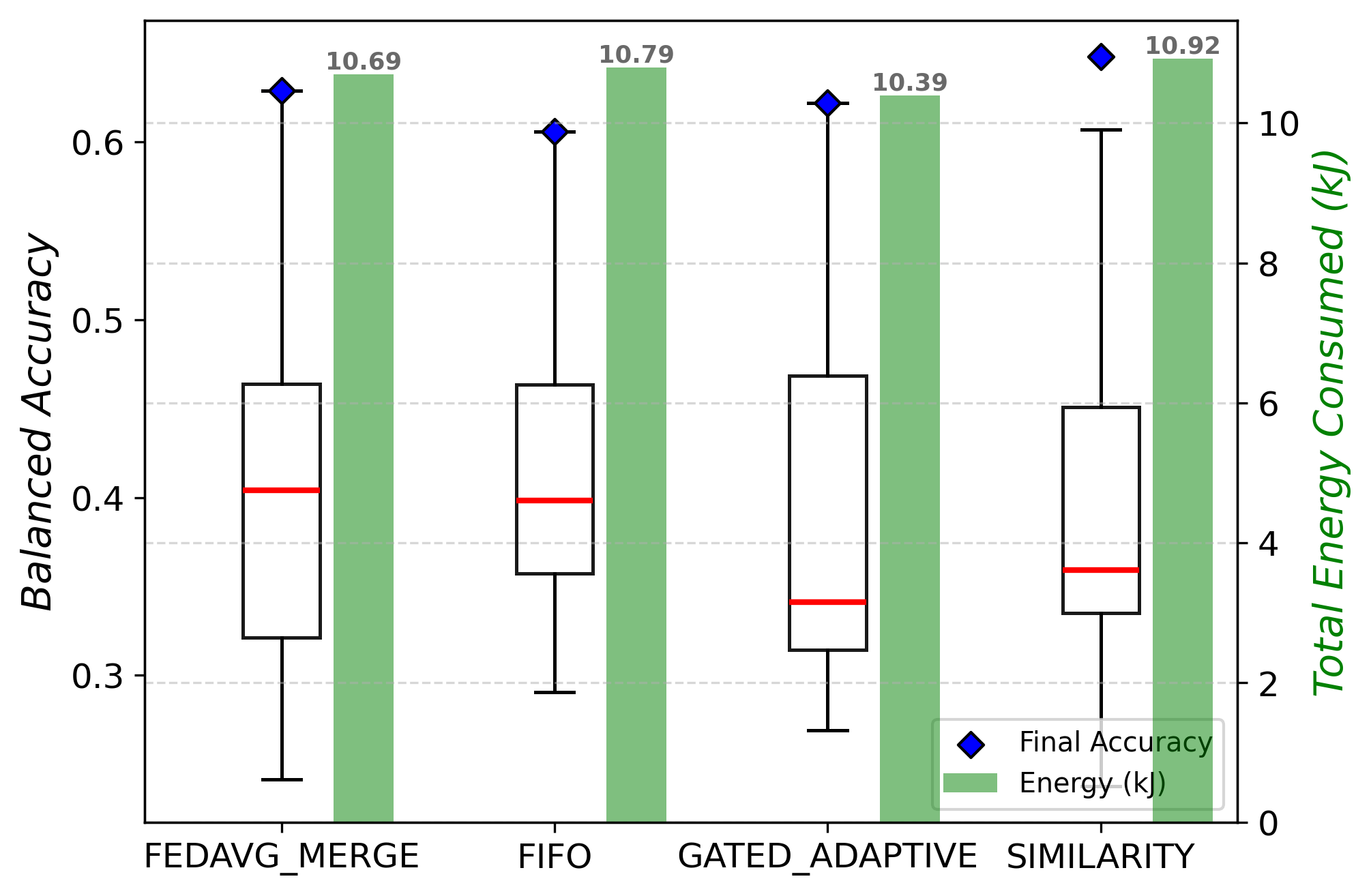}
    \caption{Buffer strategy comparison.}
    \label{fig: ablation2-strat}
\end{wrapfigure}

Figure~\ref{fig: ablation1-atten} compares the performance of \sys with and without the attention module. The ablated results indicate that the attention mechanism (with fedavg-merge buffer strategy) contributes significantly to performance gains, albeit with higher energy expenditure due to its additional computational overhead. Specifically, compared to the non-attention baseline, the median balanced accuracy improves by 15.35\% (from 0.3504 to 0.4042). Meanwhile, the final accuracy increases by 43.34\% (from 0.4386 to 0.6287), while the energy overhead increases by 4.93 kJ.

Figure~\ref{fig: ablation2-strat} presents that four buffer strategies exhibit comparable energy consumption when the attention-based feature memory is enabled, while differing in balanced accuracy. The similarity-based strategy achieves the highest final accuracy (0.6479) compared to FIFO (0.6059), gated-adaptive (0.6217), and fedavg-merge (0.6287) strategies. That is, under a fixed concept-drift type, the performance of the proposed incremental learner can be improved by incorporating a sophisticated buffer strategy.

%

\begin{figure}[!htb]
    \centering
    \subfloat[$d_h, W=15$, abrupt concept drifts.]
    { 
    \label{subfig: dim-99901}
      \includegraphics[width=.4\linewidth]{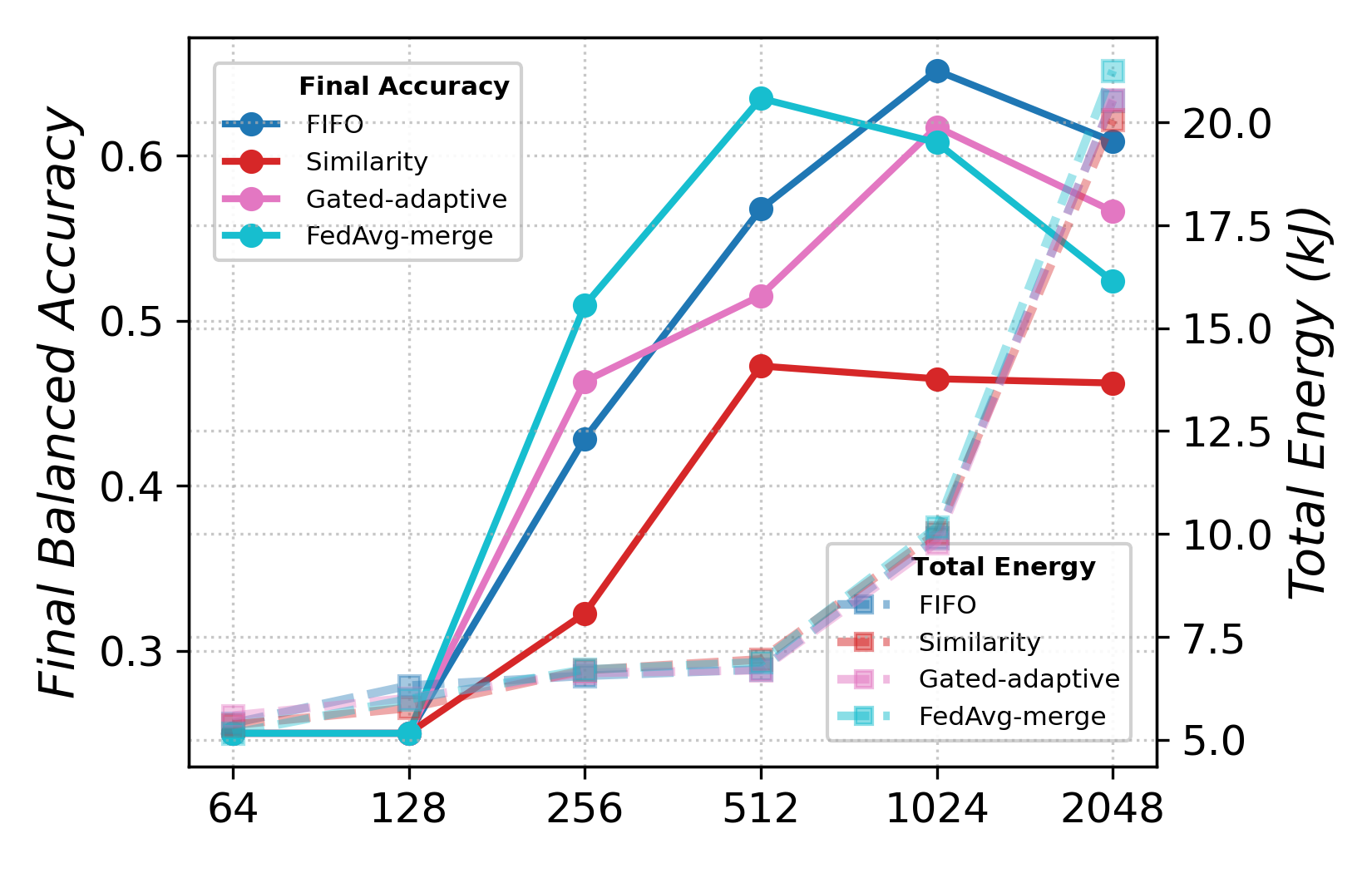}}   
    \subfloat[$W, d_h=1024$, abrupt concept drifts.]
    { 
    \label{subfig: window-99901}
      \includegraphics[width=.4\linewidth]{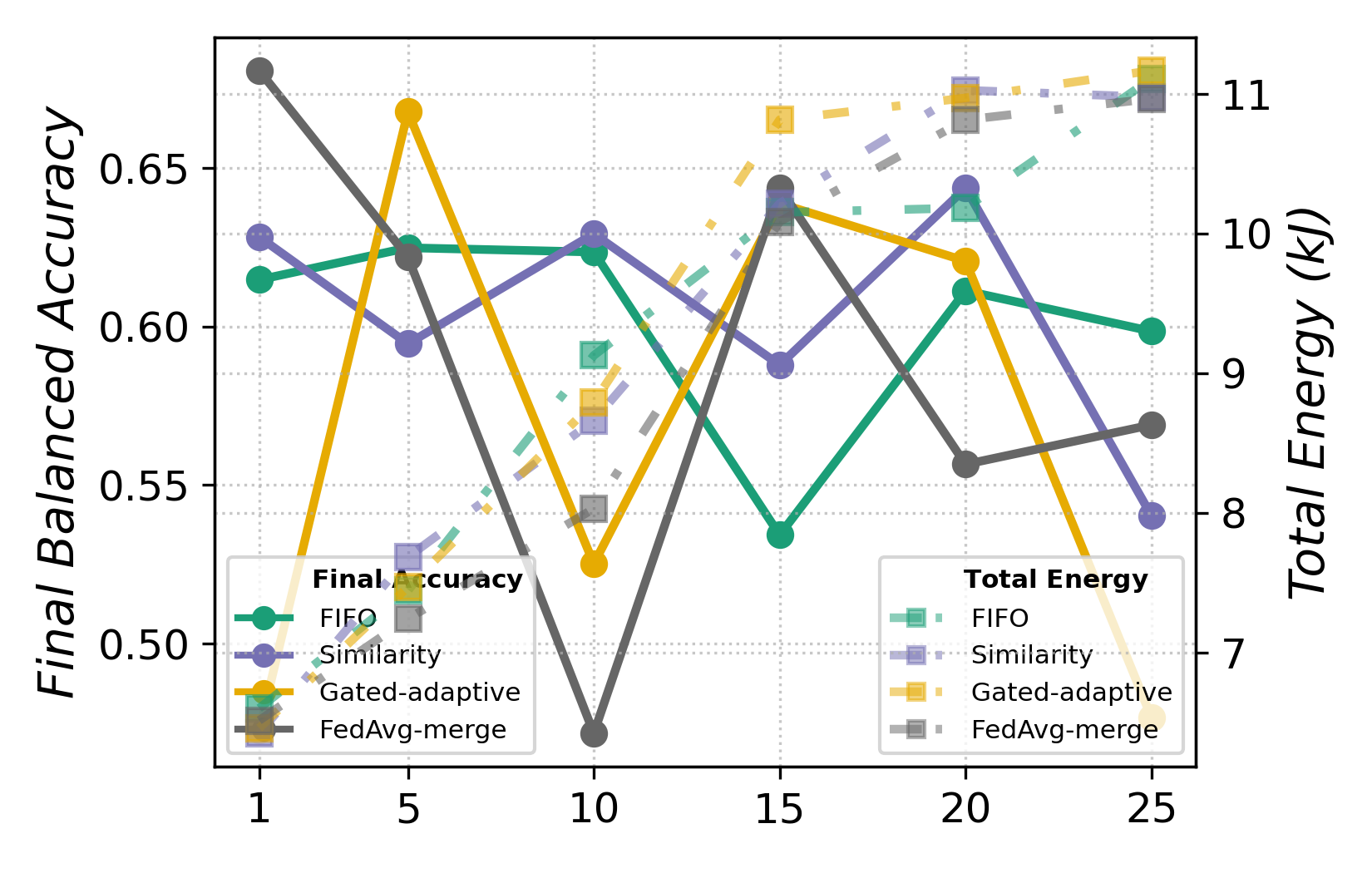}} 
    
    \subfloat[$d_h, W=15$, incremental concept drifts.]
    { 
    \label{subfig: dim-99902}
      \includegraphics[width=.4\linewidth]{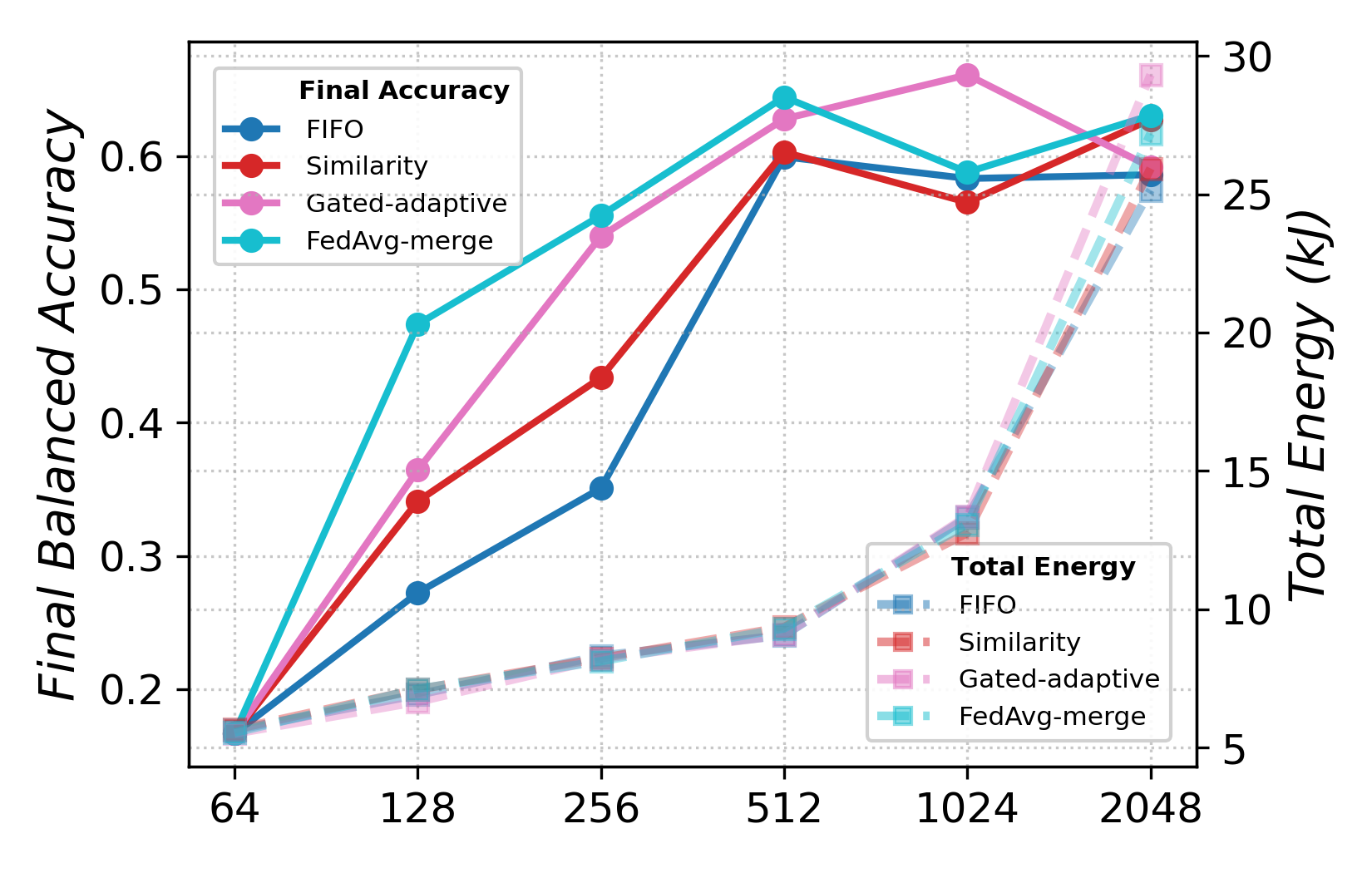}}   
    \subfloat[$W, d_h=1024$, incremental concept drifts.]
    { 
    \label{subfig: window-99902}
      \includegraphics[width=.4\linewidth]{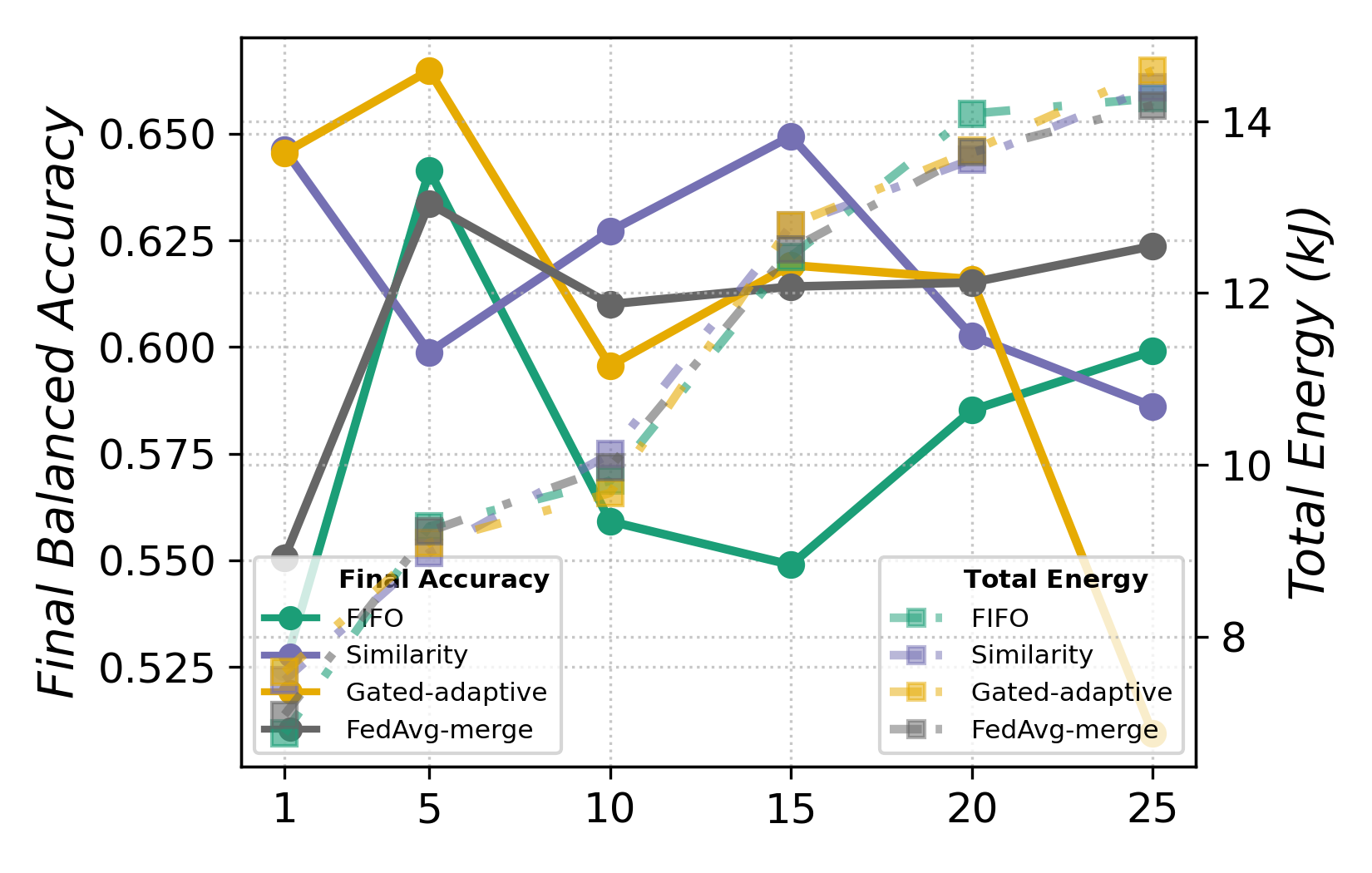}} 
    \caption{Impact of $d_h$ and $W$ under different concept drifts. }
    \label{fig: ablation2}
\end{figure}

\para{Impact of $d_h$ and $W$ under different concept drifts} We ablated the window size $W$ and hidden dimensions $d_h$ under \textit{abrupt-concept-drift} and \textit{incremental-concept-drift} datasets, as depicted in Figure~\ref{fig: ablation2}.

Figures~\ref{subfig: dim-99901} and \ref{subfig: dim-99902} compare different hidden dimension sizes under a fixed window size ($W = 15$). On one hand, the results show that increasing $d_h$ consistently improves final accuracy under both types of concept drift. Final accuracy rises steadily from $d_h = 128$ to $1024$ in the abrupt-drift setting and from $d_h = 64$ to $1024$ in the incremental-drift setting across most buffer strategies. On the other hand, under fixed $d_h$ and $W$, buffer strategy performance varies with the drift type. FIFO performs best under abrupt drifts, the gated-adaptive strategy excels under incremental drifts (at $d_h=1024$), while the similarity-based strategy performs well for incremental concept drifts but degrades under abrupt concept shifts. 

Figures~\ref{subfig: window-99901} and \ref{subfig: window-99902} show that, with a fixed $d_h = 1024$, the total energy consumption increases gradually as $W$ grows from 1 to 25. In contrast, larger window sizes do not necessarily provide more informative historical context. This is because, beyond a small number of representative model snapshots, additional buffered models tend to be redundant in representation space, especially under incremental or abrupt concept drifts.

Therefore, the \textit{attention-based feature memory} design with custom buffer strategies substantially improves the learner's predictive performance while effectively influencing the energy consumption under non-stationary distributions. 

 
\subsection{Evaluation under Different Concept Drifts}
\begin{figure*}[!htb]
    \centering 
    \subfloat[Balanced Accuracy, {abrupt} drift.]
    { 
    \label{subfig:acc_abrupt}
      \includegraphics[width=.45\linewidth]{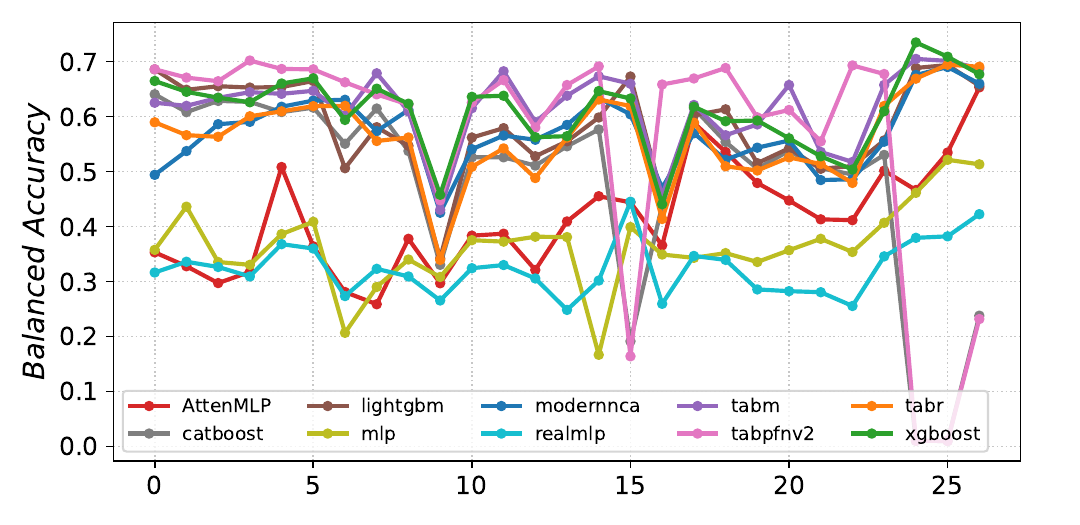}} 
    \subfloat[Energy consumed and loss, {abrupt} drift.]
    { 
    \label{subfig:energy_abrupt}
      \includegraphics[width=.45\linewidth]{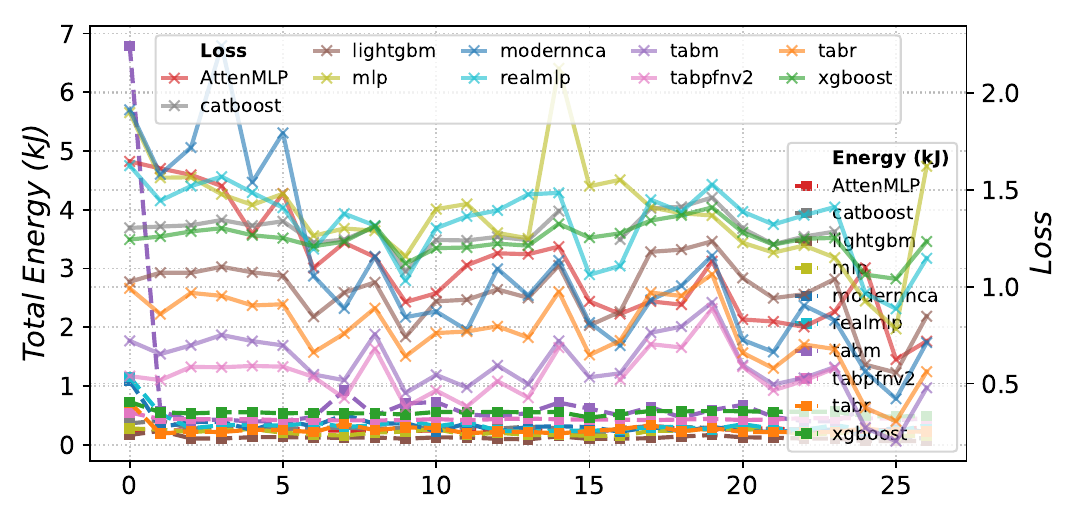}} \\
    \subfloat[Balanced Accuracy, {incremental} drift.]
    { 
    \label{subfig:acc_incremental}
      \includegraphics[width=.45\linewidth]{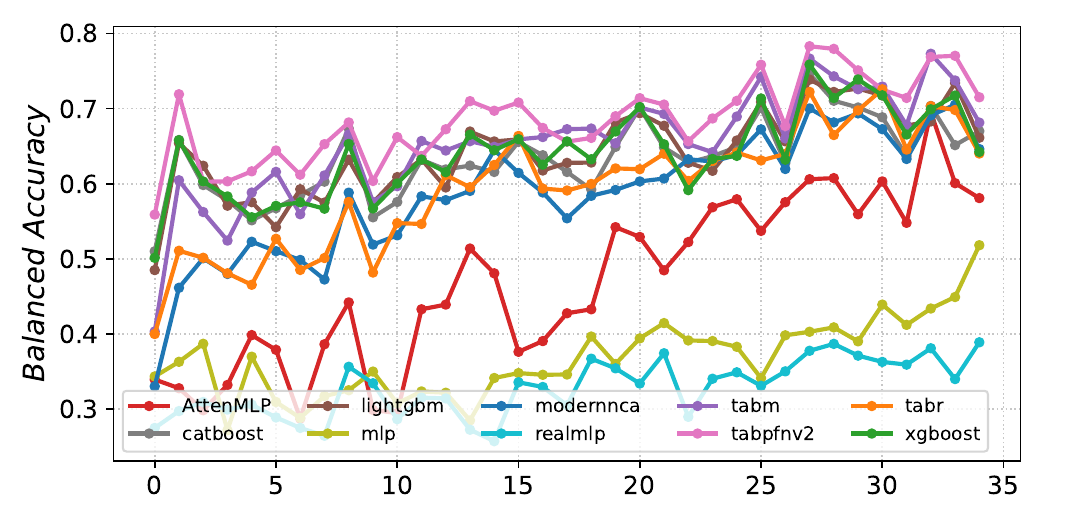}}  
    \subfloat[Energy consumed and loss, {incremental} drift.]
    { 
    \label{subfig:energy_incremental}
      \includegraphics[width=.45\linewidth]{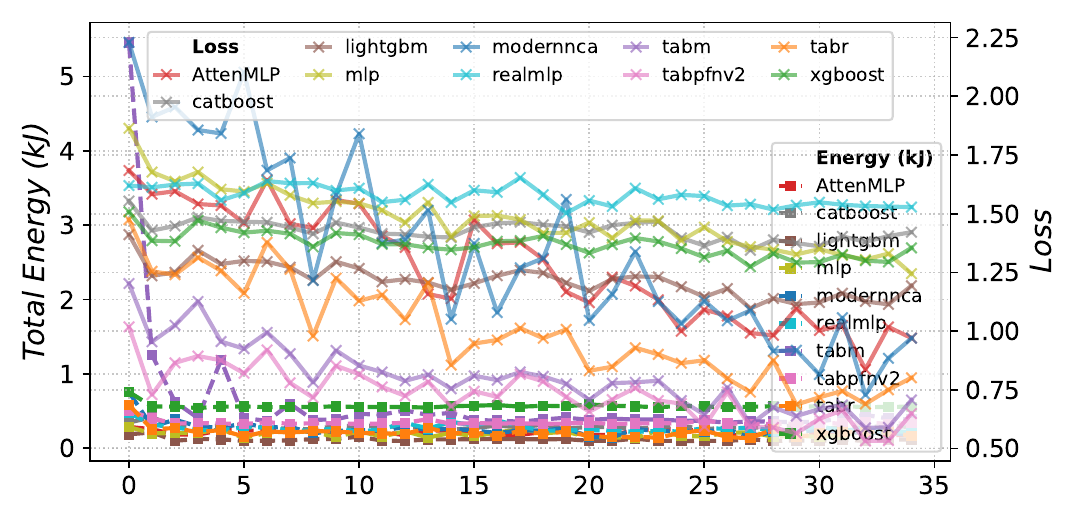}}
    \caption{Overview of the model performance and energy consumption under different concept drifts. }
    \label{fig:drifts}
\end{figure*}

Figure~\ref{fig:drifts} compares against the baseline models with the dynamic performance and energy consumed in arrival sequences, where \sys uses fedavg-merge and similarity-based buffer strategy with $d_h=1024, W=1$ for abrupt and incremental concept drifts, respectively. 

Figure~\ref{subfig:acc_abrupt} shows that CatBoost and TabPFNv2 experience a noticeable performance drop under the abrupt concept drift, while \sys (with fedavg-merge buffer strategy) adapts to the abrupt drifts. \sys experiences an immediate accuracy drop following abrupt drifts but quickly recovers in subsequent segments by leveraging latent features stored in the attention-based buffer, despite starting from a lower accuracy compared to advanced baselines. For instance, segment 16 shows a significant drop in balanced accuracy (0.3658); however, the accuracy recovers to 0.5885 at segment 17. \sys (0.6532) outperforms MLP (0.5132), RealMLP (0.4225) in the final accuracy gains, while it is comparable with TabR (0.6907), TabM (0.6874), XGBoost (0.6770), ModernCNA (0.6599), and LightGBM (0.6560). Figure~\ref{subfig:acc_incremental} shows that under incremental concept drifts, TabPFNv2 achieves the highest accuracy (0.7151), followed by TabM (0.6814), CatBoost (0.6701), LightGBM (0.6613), and \sys (0.6534), even without CL mechanisms. Equipped with the attention-based feature memory mechanism, \sys consistently outperforms RealMLP and MLP across all evaluated settings. 

Figures~\ref{subfig:energy_abrupt} and \ref{subfig:energy_incremental} present that TabM incurs the highest energy cost in the first segment under both types of concept drifts. Although LightGBM and XGBoost achieve relatively high accuracy, their loss values remain consistently higher. This reflects conservative and less well-calibrated probability estimates rather than misclassification errors. \sys and other NNs exhibit progressively lower loss values, as their use of softmax-based outputs and continuous latent representations enables better probability calibration and more confident predictions. Under both concept drift scenarios, LightGBM exhibits the lowest energy consumption across all segments. 


\begin{table}[!htb]
    \centering
    \caption{Energy-accuracy trade-off analysis.}
    \label{tab:trade-offs}
    \resizebox{.68\linewidth}{!}{%
    \begin{tabular}{llllccc}
    \toprule
    Data & Method & Final Acc & Final Loss & Energy ($J$) & Efficiency & Pareto?\\
    \midrule
    \multirow{10}{*}{\rotatebox{90}{Abrupt concept drift}} 
    &LightGBM        & 0.6560  &0.8480 & \textbf{3165.13}    & \textbf{0.7568 }    & \ding{51} \\
    &TabR            & \textbf{0.6907} & \underline{0.5618}  & 6983.89    & \underline{0.7330}     & \ding{51} \\
    &ModernNCA       & 0.6599   & 0.7119& 8813.27    & 0.6696     & \ding{55} \\
    &MLP             & 0.5132   & 1.6231& \underline{5625.17}    & 0.6419     & \ding{55} \\
    &RealMLP         & 0.4225   & 1.1474& 9140.99    & 0.5264     & \ding{55} \\
    &XGBoost         & 0.6770   & 1.2343& 14749.36   & 0.4822     & \ding{55} \\
    &CatBoost        & 0.2378   & NaN & 7279.81    & 0.4363     & \ding{55} \\
    &TabPFNv2        & 0.2317   & NaN & 11750.05   & 0.3567     & \ding{55} \\
    &TabM            & \underline{0.6874}   & \textbf{0.4786}& 20789.53   & 0.2592     & \ding{55} \\
    &\sys            & 0.6532   & 0.7195 & {6881.35}    & 0.7130     & \ding{55} \\
    \midrule
    
    \multirow{10}{*}{\rotatebox{90}{Incremental concept drift}}
    &LightGBM        & 0.6613   & 1.1930 & \textbf{4120.82}    & \textbf{0.7605}     & \ding{51} \\
    &TabR            & 0.6403   & 0.8010 & 7507.23    & \underline{0.7089}     & \ding{55} \\
    &ModernNCA       & 0.6462   & 0.9696 & 8966.41    & 0.6781     & \ding{55} \\
    &CatBoost        & 0.6701   & 1.4211 & 10163.78   & 0.6562     & \ding{51} \\
    &MLP             & 0.5183   & 1.2442 & \underline{6856.02}    & 0.6407     & \ding{55} \\
    &TabPFNv2        & \textbf{0.7151}   & \textbf{0.6486 }& 12049.59   & 0.6122     & \ding{51} \\
    &RealMLP         & 0.3891   &1.5281  & 9472.86    & 0.5136     & \ding{55} \\
    &XGBoost         & 0.6425   &1.3542  & 19844.90   & 0.2959     & \ding{55} \\
    &TabM            & \underline{0.6814}   &\underline{0.7064} & 21041.62   & 0.2579     & \ding{55} \\
    &\sys            & 0.6534   &0.9701 & 8034.49    & 0.7053     & \ding{55} \\
    \bottomrule
    \end{tabular}
    }
\end{table}
Table~\ref{tab:trade-offs} shows that under both concept drifts, \sys achieves the third-highest global efficiency, after LightGBM and TabR, although none are Pareto-optimal in the front analysis between final accuracy and total energy consumption. For abrupt drifts, \sys (0.6532, 6.88kJ) consumes 1.4\% less energy than TabR (0.6907, 7.51kJ), with a 0.0375 drop in accuracy. Under incremental drifts, \sys (0.6534, 8.03kJ) reduces energy consumption by up to 33.3\% compared to TabPFNv2 (0.7151, 12.05kJ), at a 0.062 accuracy drop. 

However, after excluding LightGBM, \sys consistently lies on the Pareto frontier and ranks among the top-performing NN baselines in terms of the energy–accuracy trade-off across both scenarios. In the abrupt drift scenario, \sys (0.7479) achieves the second-highest efficiency after TabR (0.7723), while in the incremental drift scenario, \sys attains the highest efficiency at 0.7480. TabR is ranked second at 0.7436.

These findings suggest promising directions for future work, including improved buffer management, integration of advanced model architectures, and optimized configurations to better balance energy and performance. 


%% file: chapters/S6_conclusions_v2.tex
\section{Conclusion}

This paper addresses the critical gap of EECL on tabular data streams by introducing a novel incremental MLP model called \sys. It employs a novel attention-based feature replay with context retrieval and custom buffer updates, integrated into a minibatch training loop for streaming tabular learning. 
Experiments show that \sys matches the comparable accuracy of baselines under no replay while substantially reducing energy costs. \sys achieves up to 33.3\% energy reduction compared to TabPFNv2, while sacrificing 0.062 final accuracy. Ranked third in terms of global efficiency, \sys follows LightGBM and TabR, while remaining competitive among NN–based models. When compared only against NN baselines, \sys lies on the Pareto frontier and ranks among the top-performing models with respect to the energy–accuracy trade-off. 

\para{Limitations and Future Work} Despite these exciting findings, \sys currently treats baselines on River's Insets benchmarks in an experimental setting. A promising next step is to compare the method with up-to-date models on real-life lifelong settings and broader validation on standard stream datasets, thereby enriching the benchmarks with comprehensive CL strategies. Beyond that, building a standardized energy reporting system with a hardware-level setup could help for a more robust evaluation for EECL on tabular streams. These would shed light on the influence of alternative CL strategies for SOTA baselines and EECL. 

\newpage